\documentclass{article}
\usepackage{spconf,amsmath,graphicx}
\usepackage{url}
\usepackage{booktabs} 
\usepackage{array}
\usepackage{subfig}
\usepackage{cite}
\usepackage{amsmath,amssymb,amsfonts}
\usepackage{graphicx}
\usepackage{textcomp}
\usepackage{xcolor}
\usepackage[ruled]{algorithm2e}
\usepackage{pifont}
\usepackage{tabularx,booktabs}
\usepackage{multirow}
\usepackage{makecell}
\newcommand{\etal}{\textit{et al}.}

\DeclareMathOperator*{\argmax}{arg\,max}

\def\BibTeX{{\rm B\kern-.05em{\sc i\kern-.025em b}\kern-.08em
    T\kern-.1667em\lower.7ex\hbox{E}\kern-.125emX}}
    
\title{CoughTrigger: Earbuds IMU Based Cough Detection Activator Using An Energy-efficient Sensitivity-prioritized Time Series Classifier}
\name{
    \parbox{\linewidth}{\centering
    Shibo Zhang$^{\star}$, Ebrahim Nemati$^{\dagger}$, Minh Dinh$^{\ddag}$, Nathan Folkman$^{\ddag}$, Tousif Ahmed$^{\dagger}$, Mahbubur Rahman$^{\dagger}$,Jilong Kuang$^{\dagger}$, Nabil Alshurafa$^{\star}$, Alex Gao$^{\dagger}$
    }
    }
  \address{$^{\star}$ Northwestern University, Chicago, IL, USA \\$^{\dagger}$ Samsung Research America, Mountain View, CA, USA \\$^{\ddag}$ Samsung Design Innovation Center, San Francisco, CA, USA}
\begin{document}
%
\maketitle
\begin{abstract}
Persistent coughs are a major symptom of respiratory-related diseases. 
Increasing research attention has been paid to detecting coughs using wearables, especially during the COVID-19 pandemic. 
%
Among all types of sensors utilized,
microphone is most widely used to detect coughs. However, the intense power consumption needed to process audio signals hinders continuous audio-based cough detection on battery-limited commercial wearable products, such as earbuds.
%
We present \textit{CoughTrigger}, which utilizes a lower-power sensor, an inertial measurement unit (IMU), in earbuds as a cough detection activator to trigger a higher-power sensor for audio processing and classification. It is able to run all-the-time as a standby service with minimal battery consumption and trigger the audio-based cough detection when a candidate cough is detected from IMU.
Besides, the use of IMU brings the benefit of improved specificity of cough detection. 
%
Experiments are conducted on 45 subjects and our IMU-based model achieved 0.77 AUC score under leave one subject out evaluation. We also validated its effectiveness on free-living data and through on-device implementation. 

\end{abstract}

\section{Introduction}\label{sec:intro}
Persistent coughs can be a sign of serious lung diseases, such as Chronic Obstructive Pulmonary Disease (COPD), asthma, pneumothorax, atelectasis, bronchitis, lung cancer, and COVID-19. 
Since a cough can be an important indicator of respiratory disease, reliable automated detection of coughs using everyday wearable devices is especially desirable. 
In recent decades, wearable devices such as smartphones and earbuds are becoming prevalent in our daily life.
%
%
A body of work has recently emerged, with wearable sensors showing promise in automatically detecting coughs
\cite{coughmic2011, SymDetector, listen2cough, sleepcough,hu_moment, coughwatch,increasedcough,coughbuddy,mcc}. Earbuds-based devices have been shown to be effective in cough detection~\cite{increasedcough,coughbuddy,mcc}.
%

Audio-based sensing has shown promise in detecting coughs on device, but requires higher battery consumption and introduces privacy concerns~\cite{coughmic2011, coughwatch, SymDetector, listen2cough, hu_moment, coughbuddy}. 
Battery consumption is a major concern, particularly among commercial wearables, such as earbuds, where power is a limited resource. For example, Samsung Galaxy Buds2 has a 61 mAh battery that supports up to 7.5 hours of play time per charge~\cite{buds_battery}, and adding extra features will cause a drop in battery life, and ultimately utility and user-interest. 
Some studies explored the use of an inertial measurement unit (IMU), due to its low battery footprint and computational load, to detect coughs~\cite{increasedcough, mcc}. It was soon realized that using a traditional classifier (XGBoost) to detect coughs could only yield a 47\% sensitivity and 54\% F1-Score~\cite{increasedcough}. The large number of confounding head movements made it challenging to distinguish between a cough and a non-cough.
%
%
%
\begin{figure}
\centering
\includegraphics[width=.43\textwidth]{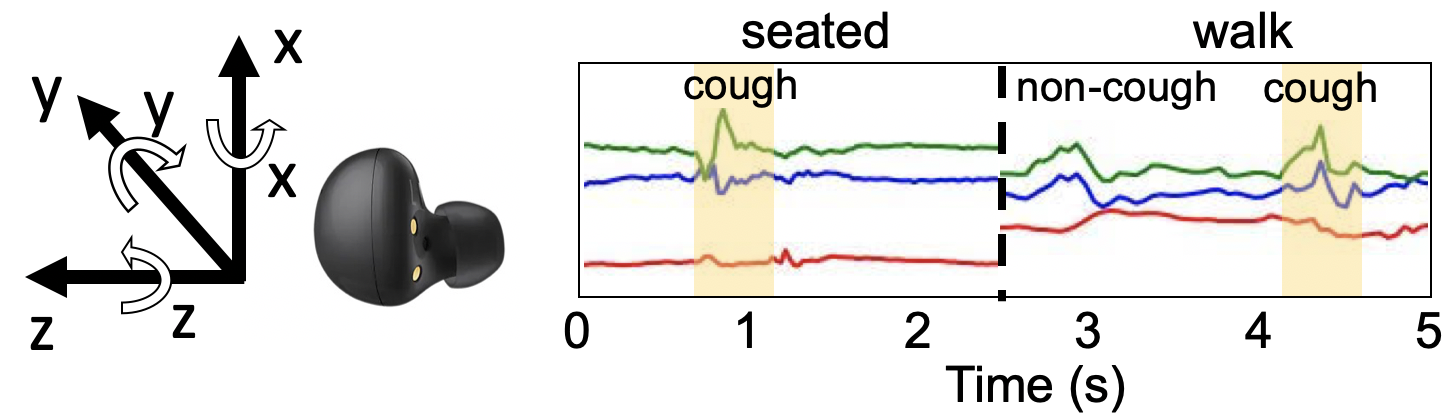}
\caption{Earbuds embedded with IMU and 3-axis accelerometer cough data (blue: x-axis, green: y-axis, red: z-axis)}
\label{fig:earbuds}
\end{figure}

Audio and IMU sensing have complementary characteristics: audio can better distinguish between coughs and non-coughs, while IMU enables battery-efficiency. Given that the majority of the time is often spent by people not coughing, the low-power IMU can be used to trigger the audio sensing pipeline when a candidate cough is detected. To do this, the IMU must yield high sensitivty, to ensure all coughs are ultimately passed onto the audio sensing. Since traditional classification methods yield low sensitivity, we propose a novel multi-center template matching algorithm to achieve high sensitivity in the IMU data. This algorithm is then used in a two-stage pipeline, where an always-on IMU triggers audio processing, only when needed, to reliably confirm the detection of a cough.

In this work, we focus on the IMU-based sensitivity-prioritized cough detection problem and propose \textit{CoughTrigger}, an IMU-based activator that utilizes a novel multi-center classifier to trigger audio processing and cough detection. 
%
\textit{CoughTrigger} can further alleviate the privacy issue because it does not require constant collecting and processing audio data. 
%
%
We summarize the contributions of this work as follows: 
\textbf{(1)} A battery-efficient dual IMU-audio cough detection framework using earbuds; we define the requirements for the first IMU stage of the pipeline and formalize it as a sensitivity-prioritized classification problem. 
\textbf{(2)} We propose \textit{CoughTrigger}, an IMU-based cough detection activator based on a novel template matching method -- Multi-Center Classifier and present superior experimental results compared with baselines. 
\textbf{(3)} We implement our proposed method on a commercial device and prove the feasibility and effectiveness of \textit{CoughTrigger} with on-device result.

\section{Methodology}\label{sec:method}




\subsection{Sensitivity-prioritized Classification}
\begin{figure}
\centering
\includegraphics[width=.15\textwidth]{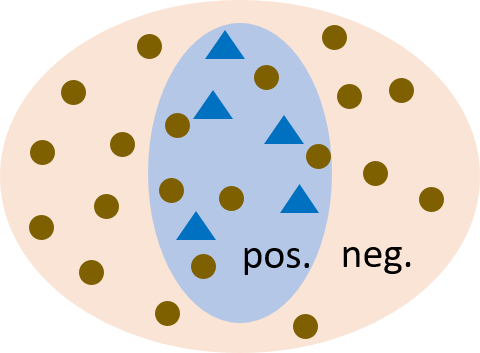}
\caption{\textcolor{black}{Illustration of sensitivity-prioritized classification}.}
\label{fig:sensitivity_clf}
\end{figure}

Unlike a traditional binary classification problem where the positive and negative samples are distinguishable 
either by human perception or machine learning techniques in a transformed feature space, 
the IMU cough data and non-cough ones are not fully separable due to an overlapping region in feature space, as illustrated in Figure~\ref{fig:sensitivity_clf}. 
To guarantee the positive samples (coughs) can be detected to trigger audio-based cough detection in the next stage, we formalize the problem as a \textit{sensitivity-prioritized} classification task, which means higher sensitivity is prioritized over specificity, and approach this problem with a novel template matching algorithm.

%
%

\subsection{Multi-Center Classifier (MCC)}

\begin{algorithm}[]
\vspace{.1cm}\SetAlgoLined
    \SetKwInOut{Input}{input}
    \SetKwInOut{Output}{output}
    \Input{Positive and negative training samples and stop criterion $H$\\}
    \Output{$K$ templates $C_k$, ($k=1, 2, \dots K$) and $K$ thresholds}
 Initialize number of clusters $K=1$; \\
 Assign all the positive samples to cluster $R_1$; \\
 Randomly select seed center $C_1$ from positive samples of cluster $R_1$; \\
 Do discrepancy-based clustering with $C_1$ to obtain total cost $L$, updated center $C_1$ and threshold;\\
 \While{total cost $L$ $>$ stop criterion $H$}{
  Select $R_t$ with the highest cost $t = \argmax_i L_i$;\\
  From the positive samples of $R_t$, select the sample which brings the largest cost increase as the new seed centroid;\\
  Do discrepancy-based clustering using $K+1$ centers $C_1, \dots, C_{K+1}$ to obtain total cost $L$, updated $K+1$ centers, and $K+1$ thresholds;\\
  Calculate cost $L_k$ for each of the $K+1$ clusters;\\
  $L$ = $\sum_{k=1}^{K+1} L_k$;\\
  $K = K + 1$;\\
 }
 \caption{Training Multi-Center Classifier}\label{alg}
\end{algorithm}
Zhang \etal~proposed a template matching algorithm~\cite{mcc}, called Multi-Centroid Classifier, which aims at iteratively creating an increasing number of clusters, each of which has its own centroid and radius and all together cover all the positive samples while include as few negative samples as possible. 
When a satisfying accuracy is achieved, the derived centroids and radii will be used in the test set as templates and thresholds to classify positive samples from negative ones.
The method has shown merits in accuracy, inference speed, and model size. 
In this work, we adopt the idea, make substantial modifications to the training phase, and propose an enhanced version, Multi-Center Classifier, on which we build \textit{CoughTrigger}. \\[3pt]
%
\textbf{Training Phase}
We present the new training procedure in Algorithm \ref{alg}. The original MCC comprises three major parts: discrepancy cost, discrepancy-based clustering, and cluster averaging. 
The discrepancy cost is used to measure and compare the purity of each cluster. 
Discrepancy-based clustering assign positive samples into clusters. 
Cluster averaging derives one template and one threshold for each cluster. 
As the original method is sensitive to regional density in the feature space, which prevents the derivation of a pseudo-optimal clustering result, we make substantial modifications to the discrepancy-based clustering and cluster averaging. Also, we change the way we select new seeds for better convergence. 
Below are the three modifications introduced:\\[3pt]
\textbf{(1)} We update the averaging step, making it easier to converge on noisy datasets. When deriving the centroid for each cluster, instead of averaging operation, we select as center the positive sample which has the minimum $Dist_{max}$ ($Dist_{max}$ is the max distance between the current sample and any other positive sample).\\[3pt]
\textbf{(2)} By modifying the clustering step, we make the algorithm less sensitive to regional density in the feature space. When clustering, we require each sample to be assigned to only one cluster instead of allowing samples assigned to multiple clusters. 
This modification brings about better convergence. We will showcase the effect with a toy dataset in Section~\ref{mcc_imp_res}.\\[3pt]
\textbf{(3)} When increasing the number of clusters, instead of adding one random centroid seed, we select the sample that brings the greatest increase to the cost function.\\[3pt]
\textbf{Inference Phase}
The inference steps are unchanged~\cite{mcc}. The distance between a test sample and each template is calculated and compared against the threshold. If the distance is smaller than the threshold, then the test sample is predicted as positive, otherwise it is negative. \\[3pt]

\vspace{-20pt}
\section{Experiment}\label{sec:res}



\subsection{Data Collection}
We generated an earbud-based cough dataset from 45 participants (15 with lung disease, 22 male, 41.4 $\pm$ 10.7 years old) with approval by the Institutional Review Board.
%
In the experiments, one earbud was worn by a participant to collect IMU data at 50 Hz and audio at 16 KHz. \\[3pt]
%
\textbf{In-lab Experiment}
To evaluate the sensitivity of our model we collected eight cough sessions, 
where we asked each participant to cough continuously for a period of time with a short pause between every two coughs. 
Cough sessions include five stationary and three non-stationary periods. Stationary periods comprised: coughing while seated (30s), coughing while lying down (30s), coughing while listening to music from an earbud (30s), coughing with background fan noise (30s), and coughing with background music/TV noise (30s). Non-stationary periods comprised: coughing while performing yoga in a quiet environment (45s), coughing while performing yoga in a noisy environment (45s), and coughing while walking (1 min). To evaluate the specificity, non-cough activities that involve signals that could resemble cough motion were collected including: eating (30s), drinking (30s), laughing (30s), scripted speech (1 min), throat clearing (30s), free head motion while talking (30s), and one additional bystander coughing session (30s). 
On average, 10.5 coughs are captured in each cough session for each participant. 
We aim at detecting each single cough in every session while preventing false alarms.\\[3pt]
\textbf{Free-living Experiment}
Enrolled participants also took part in a free-living experiment, where they were asked to cough naturally in the morning and in the afternoon for one week. At each time, data were collected for coughs while seated for 30s and while walking for 30s.\\[3pt]
%
%
%
%
\textbf{Annotation}
We annotated every single cough using audio. 
Due to the limitations of data transmission from earbud to data logging app, there is a random time drift of up to about 400 ms between the audio and IMU data. 
%
We resolved the asynchronization by observing the data and identifying peaks in the  x-axis of accelerometer, which corresponds to the most probable motion of the wearer while coughing. 
We labeled 45 participants for the in-lab experiment and the 15 participants from the lung disease cohort in the free-living experiment.

\subsection{Model Development}
%
\textbf{Data Preprocessing}
3-axis accelerometer data are preprocessed using a moving average filter with a window size as 10 samples and a Butterworth high-pass filter ($\omega_c=3\pi$). 
%
Positive samples are extracted from seated and lying down cough sessions using a 0.4s window centered around the annotated IMU cough moment. 
To increase the variety of cough data during training, we applied three time series augmentation methods, namely jittering, scaling, and magnitude warping~\cite{data_aug} on cough data enlarging the positive sample size by four times. 
For negative samples, we segmented the non-cough session accelerometer data using a 0.4s sliding window with 0.1s stride size. 
Then we randomly subsampled four times the positive samples from non-cough sessions as negative samples, in order to balance class ratio.
\\[3pt]
\textbf{Model Training}
%
To expedite the training process, 
we trained one MCC model with stop precision criterion as 0.8 for each participant in the training set using multi-variate DTW distance measure, then aggregated the models of each training participant together and ran on all the training participants. 
We used a greedy algorithm to rank the templates and select the top $K$ templates based on template importance measured by how many new positive samples are hit by each template. 
\\[3pt]
\textbf{Model Testing}
We applied MCC with $K$ templates on the accelerometer data of the test participant's cough and non-cough sessions, with 0.4s window size and 0.02s stride size. 
Then, we aggregated and merged all the predicted cough windows to determine the final cough event prediction. 
In our case, the capability of manually adjusting the trade-off between sensitivity and specificity is desirable. 
A benefit of MCC is that we have two ways to adjust it: to choose the number of top $K$ templates used and to adjust the thresholds of templates. When more (less) templates are used, or when we increase (decrease) the thresholds of templates, it leans towards a higher sensitivity (specificity). 
\\[3pt]
\textbf{Evaluation Method}
Leave one subject out cross validation (LOSOCV) is adopted.
%
For cough sessions, we compared the predicted cough segments against the ground truth coughs. 
True positives are defined as predicted cough segments intersected by ground truth coughs. We calculated the sensitivity of cough sessions as a ratio of true positives to ground truth coughs. 
We used sample-level specificity to test how well it can specify non-cough events. 
%
The model is validated on free-living data in a similar manner. 
\section{Result}
\subsection{Improvement of MCC}\label{mcc_imp_res}
\begin{figure}
\centering
\includegraphics[width=.35\textwidth]{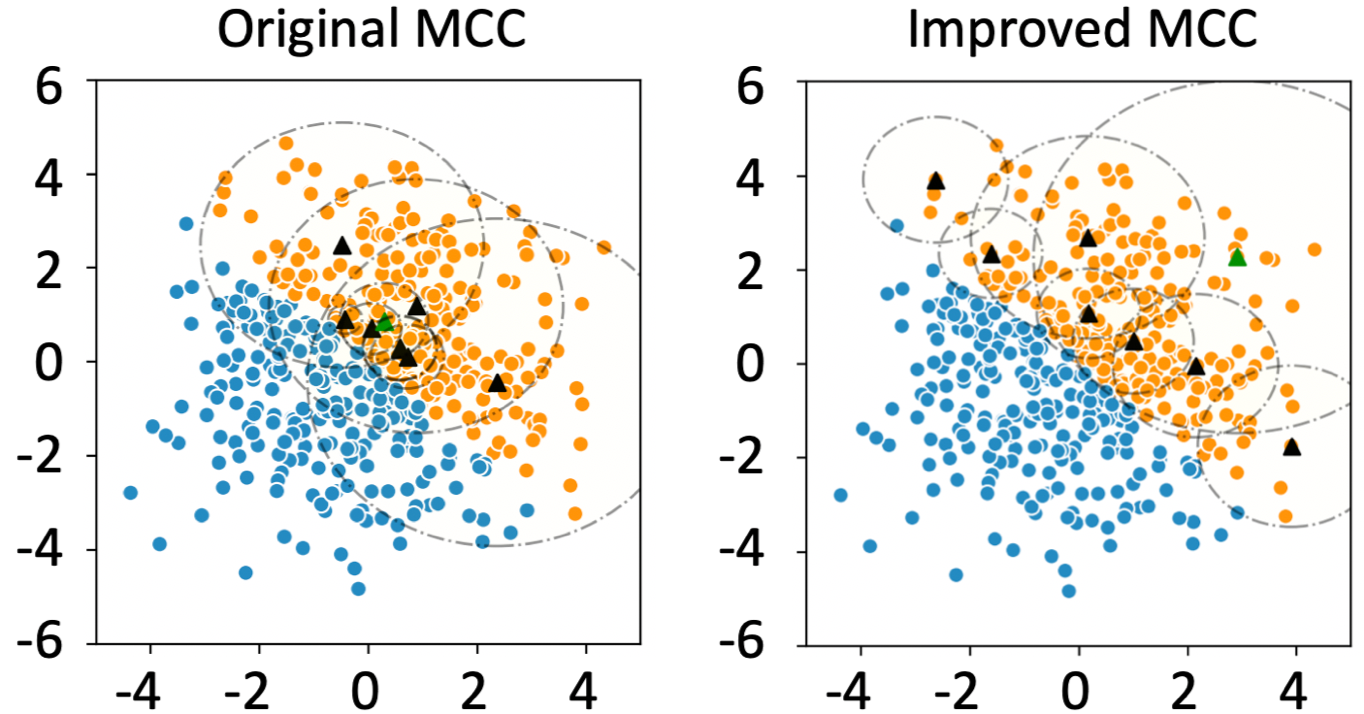}
\caption{Illustration of Multi-Center Classifier improvement.}
\label{fig:mcc_imp}
\end{figure}

We used a synthetic dataset generated from Gaussian distribution to validate the improved MCC, as in Fig.~\ref{fig:mcc_imp}.
The original MCC achieved 0.57 test accuracy with 10 centroids. 
Using improved MCC, 0.99 test accuracy was obtained with 8 centers. 
The first reason is that the original method allows one sample to belong to multiple clusters, increasing the possibility of highly overlapped clusters. 
Instead, the new method can separate clusters towards different directions. 
Second, the new clustering averaging method makes MCC less sensitive to regional density.

\vspace{-6pt}

\subsection{\textit{CoughTrigger} Results}
\begin{figure}
\centering
\includegraphics[width=.48\textwidth]{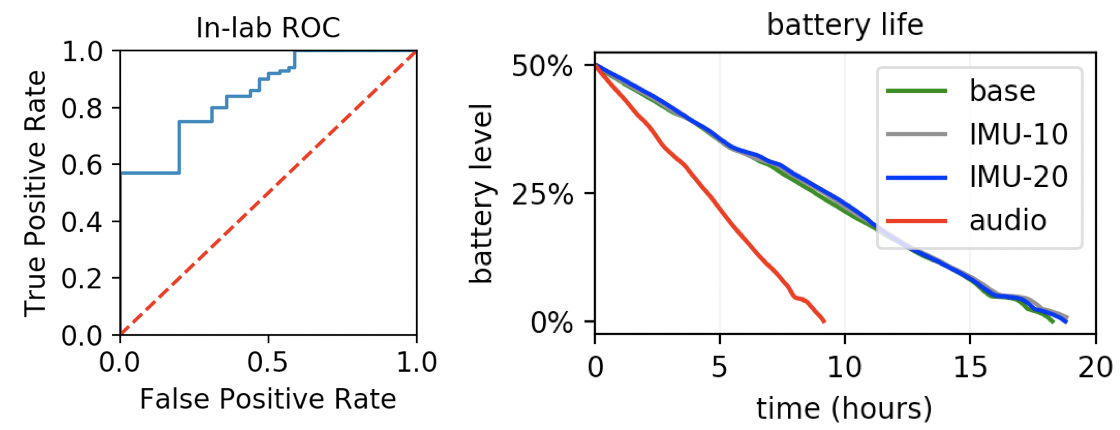}
\caption{Left: In-lab experiment ROC curve; Right: Battery life comparison.
}
\label{fig:battery}
\end{figure}

\textbf{Choice of Input Data}
We compared all the combinations of the 3-axis accelerometer data, including three using one axis, three using two axes, and one using three axes. 
The best result is achieved with both the x- and y-axis. That aligns with the IMU direction in Fig.~\ref{fig:earbuds} as z-axis only contributes to sensing motion in the left/right direction, which is not as useful in detecting coughs. 
We tried with only 3-axis gyroscope and combining both accelerometer and gyroscope data but observed lower accuracy. 
\\[3pt]
%
%
%
\textbf{In-lab Experiment Result}
Under LOSOCV, when training MCC, on average 14.2 templates were generated for each participant in each fold. When using top five templates, we achieved 90\% sensitivity on average for the five stationary cough sessions, and an average sensitivity of 86\% on all cough sessions. 
The average specificity is 53\% across all non-cough sessions. 
When the number of adopted templates ranges from 1 to 30, we achieved 0.77 AUC with all the 15 sessions.
We present the ROC curve of \textit{CoughTrigger} in Fig.~\ref{fig:battery} Left. 
We emphasize that our specificity evaluation was designed for the worst case scenario with a variety of activities. In real life, as for most of the time the wearer is stationary, the specificity is expected to be higher.
%
%
%
\\[3pt]
%
%
%
\textbf{Free-living Experiment Result}
We trained with the in-lab data from all 45 participants and tested on the free-living data from the 15 participants in the disease cohort. When we applied top 10 templates, we received an average sensitivity of 87\% during the stationary periods, and an average sensitivity of 82\% on both seated and walking cough periods. Although we only have cough sessions in the free-living setting, we found a 55\% average specificity across both sessions for all the participants. When we adjusted the thresholds of each templates, we received 0.80 AUC for stationary coughs and 0.73 AUC for stationary and walking sessions. 
\\[3pt]
%
%
%
%
\textbf{Baseline Methods}
Since there is no existing out-of-box sensitivity-prioritized classification method, we investigated ways that may apply, including one-class classifier and adjusting the decision boundary of a traditional classifier. Because one-class classifier only utilizes the positive class and models the distribution of positive samples, the overlapping of two classes should not interfere with modeling of the positive class.
We tested the popular OC-SVM 
which relies on identifying the smallest hypersphere consisting of all the positive samples~\cite{ocsvm,ocsvm1}.
%
%
We utilized the same preprocessing steps and concatenated x- and y-axis into one vector of size 40 as input.
After adjusting the hyperparameters in a large range with different kernels (linear, RBF, and Sigmoid), we only received 0.51 AUC, which is no better than random guess.
%
For a traditional classifier, we implemented a 3-layer NN with 20, 10, and 5 neurons in each layer and we observed the same result as OC-SVM. 
\\[3pt]
%
%
%
%
\textbf{On-device Implementation}
We implemented \textit{CoughTrigger} on Samsung Galaxy Buds2. Fig.~\ref{fig:battery} Right shows the battery life of base firmware without cough detection, \textit{CoughTrigger} using 10 and 20 templates integrated in the base firmware, and an integrated audio-based cough detection method~\cite{coughbuddy}.
We see that the base firmware without cough detection has around 18 hours battery life, and integrating \textit{CoughTrigger} (10 or 20 templates) makes no significant change in the battery life. The IMU-based cough detection implementation consumes less battery than an audio-based method, which shows the feasibility of leveraging \textit{CoughTrigger} as a first-stage filter to reduce battery consumption.
%



\section{Conclusion}\label{sec:conc}


In this work, we introduce a battery-efficient earbuds-based two-stage IMU-audio cough detection framework and formalize the IMU first stage as a sensitivity-prioritized classification problem. 
We propose using a novel multi-center classifier as a first-stage cough detection activator. By conducting in-lab, free-living, and on-device experiments, we demonstrate the feasibility and effectiveness of our proposed method. 
Our proposed sensitivity-prioritized template matching algorithm can be adopted as a plug-and-play module for other sensor-fusion wearable applications.

\bibliographystyle{IEEEbib.bst}
\bibliography{paper, refs}

\end{document}